\documentclass[conference]{IEEEtran}
\IEEEoverridecommandlockouts
\usepackage{cite}
\usepackage{amsmath,amssymb,amsfonts}
\usepackage{algorithmic}
\usepackage{graphicx}
\usepackage{textcomp}
\usepackage{xcolor}
\usepackage{enumitem}
\usepackage{calc}
\usepackage{flushend}
\def\BibTeX{{\rm B\kern-.05em{\sc i\kern-.025em b}\kern-.08em
    T\kern-.1667em\lower.7ex\hbox{E}\kern-.125emX}}

\newcommand{\sub}[1]{\allowbreak\texttt{/d/#1}}
\usepackage{booktabs}
\usepackage{multirow}
\definecolor{lightblue}{RGB}{230,242,252}
\usepackage[table]{xcolor}
\usepackage{soul}
\usepackage{enumitem}
\usepackage{array} 

\begin{document}

\title{\textit{All the Pieces Matter!} A Six-stage Analysis of an Underground Fraud Ecosystem on Dread}

\author{Submission to eCrime 2026, \#10}

\author{
\IEEEauthorblockN{
\textbf{Megha Sundriyal\textsuperscript{1},
Dylan Raharja\textsuperscript{2},
Rui Cao\textsuperscript{2},
Roy Ka-Wei Lee\textsuperscript{3},
Meeyoung Cha\textsuperscript{1}}
}
\IEEEauthorblockA{
\textsuperscript{1}Max Planck Institute for Security and Privacy, Germany
\textsuperscript{2}Singapore University of Technology and Design, Singapore \\
\textsuperscript{3}University of British Columbia, Canada
}

\IEEEauthorblockA{
\textit{\{megha.sundriyal, mia.cha\}@mpi-sp.org,
\{dylan\_raharja, rui\_cao\}@sutd.edu.sg, 
roy.lee@ubc.ca}
}
}

\maketitle

\thispagestyle{plain}
\pagestyle{plain}

\begin{abstract}
Cyber-enabled fraud is a dominant global threat that causes severe financial loss, identity theft, and widespread psychological harm. This illicit economy relies on a highly specialized underground ecosystem where distinct actors exchange technical tools, operational techniques, and laundering channels. This research presents a case study of one such live fraud ecosystem from the perspective of a cohesive six-stage value chain spanning from preparation to monetization: Capability, Data Acquisition, Technique Sharing, Identity Forgery, Execution, and Extraction. Analyzing 67,976 posts and 305,048 comments from nine communities on the Tor-based darknet forum Dread, we map user trajectories, content distribution, and the adoption of generative AI. Our findings show that structural cohesion within the ecosystem is maintained by a thin layer of `bridgers' who generate the majority of all content. Against a degree-preserving null model, three knowledge-building upstream stages of the value chain form a tightly coupled cluster, whereas the remaining three operational downstream stages resolve into highly specialized populations with small personnel overlap. Finally, we demonstrate that generative AI is being rapidly but unevenly absorbed across the value chain; the technology saturates knowledge-building communities while operational stages remain largely untouched. By quantifying criminal specialization and structural integration, our network-based approach provides a framework for targeted, high-impact law enforcement interventions.
\end{abstract}


\section{Introduction}
\label{sec:introduction}

\noindent
Consumer-facing fraud has escalated into one of the most pervasive and destabilizing dimensions of modern cybercrime. In the United States, annual consumer losses reached \$12.5 billion in 2024~\cite{ftc_2024,fraud_2024}, while broader cyber-enabled crime accounted for \$16.6 billion in losses over the same period \cite{fbi_2024}. This threat landscape is undergoing a profound structural shift, heavily accelerated by the rapid integration of artificial intelligence (AI). 
In 2025, the FBI logged over 22,000 formal complaints involving AI-enabled fraud, with reported losses exceeding \$893 million \cite{ic3_2025}. These reports suggest a fundamental evolution in criminal architecture: modern fraud schemes have transitioned from the purview of isolated offenders to highly distributed, specialized networks. In particular, detection has become far challenging because the technical tools, operational strategies, and financial laundering channels required for successful exploitation are independently engineered and traded across disparate illicit markets, converging only at the precise moment of execution.


The infrastructural backbone of this ecosystem is increasingly exposed by targeted state interventions. Over the past five years, law enforcement disruptions of underground forums and marketplaces have consistently revealed these platforms to be critical coordination hubs for global fraud. For example, the 2023 seizure of Genesis Market dismantled an operation that facilitated account takeovers and bank fraud by commercializing approximately 1.5 million compromised digital identities \cite{genesis}. Similarly, the 2022 takedown of RaidForums and the rapid emergence of its successor, BreachForums, demonstrated how illicit credential markets fuel industrial-scale credential-stuffing and impersonation campaigns \cite{raid_forums}. These cases share a common pattern: while profitable fraud occurs downstream (i.e., at the final stage of execution and monetization), the vital coordination that enables it happens upstream (i.e., during the backend preparation and supply stage). This coordination is concentrated within forums, where skills, stolen data, and fraud techniques are systematically exchanged as both products and knowledge.

Prior research has extensively characterized individual components of this cybercrime infrastructure. Empirical measurements have traced the longitudinal expansion of anonymous marketplaces \cite{soska2015measuring}, mapped the illicit data trade \cite{holt2013exploring}, and documented the broader industrialization of cybercrime as an enterprise \cite{lusthaus2018industry}. Complementing these, crime-script analyses have decomposed specific fraud sub-processes, most notably stolen-data market \cite{hutchings2015crime}, into sequential stages of offender activity \cite{cornish1994procedural}, reflecting the intrusion-side decomposition of the cyber kill chain \cite{hutchins2011intelligence}. However, existing literature treats these components in isolation. The underground fraud ecosystem has not yet been analyzed as a globally \emph{coupled} network of functionally specialized communities, each liable for a distinct phase of the fraud lifecycle and interconnected through personnel, knowledge, and discourse. Without such a systemic perspective, critical questions regarding network vulnerabilities remain unanswerable. These include the extent to which the fraud chain is integrated, which communities act as proxies between otherwise weakly coupled sub-ecosystems, and which players connect distinct stages of the fraud value chain.


We address this systemic gap by analyzing Dread, a prominent Tor-based discussion platform serving as a primary hub for contemporary cybercriminal collaboration. Utilizing a curated corpus of 67,976 posts and 305,048 comments across nine fraud-oriented communities (`subdreads'), we formalize unstructured text into an interconnected six-stage value chain. The resulting framework shifts the analytical paradigm from tracking isolated threat actors to quantifying the global structural integration of the ecosystem as a whole. Concretely, our investigation is guided by three research questions:

\vspace*{1mm}
\begin{enumerate}[
    label=\textbf{RQ\arabic*:},
    leftmargin=*,
    align=left]
    \item \textit{To what extent do users span the stages of the fraud value chain, and what does this reveal about criminal specialization?}

    \item \textit{How are content production and user engagement distributed across the ecosystem, and what does the presence of structural asymmetry reveal about the operational logic of these communities?}

    \item \textit{Where along the cybercrime value chain is generative AI being adopted, and where are these capabilities structurally concentrated?} 
\end{enumerate}


Ultimately, our work attempts to move beyond a descriptive catalog of underground activities; by identifying the structural channels and cross-stage 'bridging' mechanisms that sustain this illicit economy, we provide the empirical blueprint necessary to transition from reactive defense to targeted, system-wide disruptions.

\section{Background}

\noindent
Research on cyber-enabled fraud and underground economies spans studies of cybercrime markets, procedural analyses of fraud operations, and work on the organizational structure of online criminal ecosystems. 

A substantial body of literature has analyzed underground marketplaces and forums as financial establishments. Early measurement studies portrayed the lifecycle and resilience of anonymous marketplaces, emphasizing their role as infrastructures for trust, reputation, and illicit trade \cite{soska2015measuring}. Other work explored the trade in stolen credentials, payment data, and compromised accounts, documenting how specialization and brokerage emerge in cybercriminal markets \cite{holt2013exploring}. Studies of carding communities further revealed that underground forums support reputation-building, operational coordination, and fraud monetization pipelines \cite{haslebacher2017all}. Research on the broader industrialization of cybercrime has similarly asserted that many forms of online offending now resemble legitimate digital industries, with differentiated labor, service provision, and transnational supply chains \cite{lusthaus2018industry}. Related studies have analyzed forums as social spaces where offenders exchange techniques, recruit collaborators, and develop criminal expertise \cite{holt2010examining}. Recent work has examined communication practices and platform migration among underground forum users, emphasizing how offenders adapt their coordination strategies across communication infrastructures \cite{hoheisel2024dark}.

A second line of work approaches fraud through procedural or script-based analysis. Crime script theory conceptualizes offenses as sequences of coordinated actions carried out across multiple stages \cite{cornish1994procedural}. This was applied to cybercrime and fraud ecosystems to decompose activities such as carding, phishing, and the monetization of stolen data into operational sub-processes \cite{hutchings2015crime}. In parallel, cybersecurity research introduced the cyber kill chain as a staged model of intrusion activity \cite{hutchins2011intelligence}, later extended in adversarial workflows and criminal service ecosystems. These perspectives emphasize that successful fraud depends not on a single action but on the orchestration of capabilities, identities, infrastructure, and monetization channels over time. Recent empirical work on ransomware ecosystems and cybercrime monetization has reinforced this view by modeling the economic dependencies between attackers, brokers, and victims \cite{meurs2023ransomware,cable2024showing}.

Recent literature highlights increasing modularity and specialization across online criminal ecosystems. Studies of ransomware ecosystems describe increasingly modular structures in which malware developers, initial-access brokers, negotiators, and money launderers operate as distinct but interdependent actors \cite{paquet2019ransomware}. Similar patterns have been identified in phishing-as-a-service operations and credential markets, where offenders rely on outsourced technical and operational support \cite{huang2015framing}. Network analyzes of cybercriminal forums suggest that a relatively small set of highly connected actors often facilitates information diffusion and cross-community interaction \cite{dupont2016ecology}. Other work has focused specifically on identifying expert or influential actors within cybercrime forums based on technical specialization and participation behavior \cite{ruellan2024identifying}. Research also expanded into emerging fraud domains, including cryptocurrency investment scams, which reveal increasingly sophisticated and globally distributed fraud infrastructures \cite{ordekian2024sinister}.

A parallel body of research has examined the effectiveness and unintended consequences of cybercrime interventions. Studies have explored how law-enforcement takedowns, moderation actions, and platform disruptions affect criminal ecosystems and offender adaptation \cite{meurs2024against}. Related work has emphasized that cybersecurity interventions can produce unintended harms or displacement effects, particularly when interventions alter offender behavior without reducing underlying incentives \cite{chua2019identifying}. These findings underscore the importance of understanding ecosystem structure before designing disruption strategies. Additional work has also examined ownership, control, and abuse within domain ecosystems used to support cybercrime operations \cite{erfan2024owned}.

Despite these advances, prior work has generally focused on either individual marketplaces, specific fraud techniques, or isolated offender communities. Comparatively little attention has been paid to the \emph{connective structure} linking the different functional stages of the fraud ecosystem. Existing studies rarely examine whether the same users operate across multiple stages, who act as bridges between otherwise disconnected sub-ecosystems, or how knowledge and participation flow across the broader fraud supply chain. In practice, however, contemporary fraud operations depend on these cross-stage connections: technical capabilities, stolen identities, and laundering mechanisms are frequently sourced from different communities and recombined dynamically during execution.

Our work contributes to this literature by modeling fraud-oriented communities as components of a coupled ecosystem rather than as isolated markets or discussion spaces. Using longitudinal data from Dread, we examine specialization and overlap across communities, identify users who bridge multiple operational stages, and analyze how emerging technologies, such as AI, diffuse through the ecosystem. In doing so, we provide an ecosystem-level account of fraud coordination that complements prior work on underground markets, cybercrime services, and procedural models of offending.

\section{Dataset}
\label{sec:dataset}

\noindent
Studying the underground fraud ecosystem at scale requires a venue where fraud knowledge is openly produced, contested, and circulated rather than privately transacted. We therefore turn to Dread, one of the largest darkweb discussion forums. 

\subsection{Platform} 
\label{sec:platform}  

\noindent
We study Dread, a Tor-based discussion forum launched in 2018 and currently among the largest English-language darknet forums \cite{Wikipedia_dread}. Dread is structured as a federation of topic-scoped \emph{subdreads} (also called communities), each moderated independently and focused on a specific theme, ranging from operational security and harm reduction to marketplace discussion, and with a significant set of communities focused on the production and circulation of fraud knowledge. The platform's structural and cultural conventions are modeled on Reddit, where users submit \emph{posts} that receive threaded \emph{comments}, and community references are typically written in the form \texttt{/d/subdread\_name}. Unlike marketplaces, Dread does not host transactions directly. Rather, it serves as the substrate around them, making it particularly well-suited for studying how illicit knowledge is produced, discussed, and circulated.

\subsection{Community Selection}
\label{sec:selection}

\noindent
Since no authoritative taxonomy of fraud-relevant subdreads exists. We therefore constructed one via single-hop snowball sampling \cite{goodman1961snowball} from a single seed, expanding it through observed cross-community references. 

\subsubsection{Seed} We seeded the sample with \sub{Fraud}. Two properties motivated this choice. First, \sub{Fraud} is among the highest-volume and most consistently active fraud-themed communities on Dread. Second, its declared scope is explicitly restricted to IT-enabled financial fraud, in contrast to larger but more diffuse subdreads whose discussions range across the full spectrum of darknet activity. \sub{Fraud} therefore serves as a high-recall entry point into the platform's fraud-oriented portion without admitting unrelated illicit content.

\subsubsection{Expansion} From the seed, we identified outbound references as direct textual mentions of the form \texttt{/d/X} (case-sensitive) in \sub{Fraud}'s posts and comments. Each subdread was ranked by the number of times it was mentioned. We retained the 20 most-referenced subdreads. No ties occurred at the cutoff. From these 20 candidates, we manually filtered communities on three principled grounds.
\begin{itemize}
    \item \textit{Forum-wide discussion hubs:} We excluded subdreads whose scope spans the entire Dread population rather than the fraud ecosystem specifically. These include \sub{OpSec}, \sub{Jobs4Crypto}, and \sub{DarkNetMarkets}, which serve as cross-cutting venues that draw participants from across the platform's full range of themes. Their content overlaps with fraud topics, but including them would skew community-level measurements and conceal the internal structure of the fraud ecosystem. 
    
    \item \textit{Marketplace channels:} We excluded subdreads that function as the official discussion and customer-support venues of specific commerce platforms, including \sub{WorldMarket} and \sub{Nexus}. These are channels where illicit goods are transacted, but they are not communities in which fraud techniques are produced, debated, and circulated as knowledge. The distinction between \emph{transactional venues} and \emph{knowledge-producing communities} is central to our focus of study. 
    
    \item \textit{Low-activity communities:} Finally, we dropped subdreads with minimal levels of interaction, removing communities with fewer than 100 posts. Such sparse activity provides insufficient behavioral and structural signals for reliable community-level statistical analysis, including measures of participation, topic prevalence, and user overlap. 
\end{itemize}

\subsubsection{Final set} After applying the exclusion criteria, the procedure yields a final set of nine communities: the seed \sub{Fraud} and eight additional subdreads. We list them, together with their sizes, in Table~\ref{tab:dataset_overview}.

\begin{table}[t]
\caption{Overview of the nine subdreads retained for analysis along with posts, comments, and the unique users count. \textsuperscript{\dag}Total user counts are non-additive due to overlap across subdreads.
}
\label{tab:dataset_overview}
\resizebox{\columnwidth}{!}{
\begin{tabular}{llrrr}
\toprule
\textbf{Stage} & \textbf{Subdread} & \textbf{Posts} & \textbf{Comments} & \textbf{Users} \\
\midrule

Capability & \sub{hacking} & 11,814 & 53,518 & 15,025 \\

\multirow{2}{*}{Acquisition} & \sub{Carding}  & 17,282 & 82,493 & 19,495 \\
                             & \sub{skimmers} & 408 & 1,876 & 605 \\

\multirow{2}{*}{Technique} & \sub{Fraud} & 31,133 & 137,126 & 25,263 \\
                            & \sub{Refunds} & 617 & 2,414 & 770 \\

\multirow{2}{*}{Identity} & \sub{FakeID} & 2,121 & 6,712 & 3,276 \\
                          & \sub{Forgeries} & 1,499 & 6,313 & 2,479 \\

Execution  & \sub{Banks}  & 1,476 & 4,991 & 2,065 \\

Extraction & \sub{Laundromat} & 1,626 & 9,605 & 3,478 \\
\midrule

\textbf{Total} & \textbf{9 subdreads}& \textbf{67,976}  & \textbf{305,048}  & \textbf{50,761\textsuperscript{\dag}} \\
\bottomrule
\end{tabular}}
\end{table}

\begin{figure*}[t!]
\centering
\includegraphics[width=\textwidth]{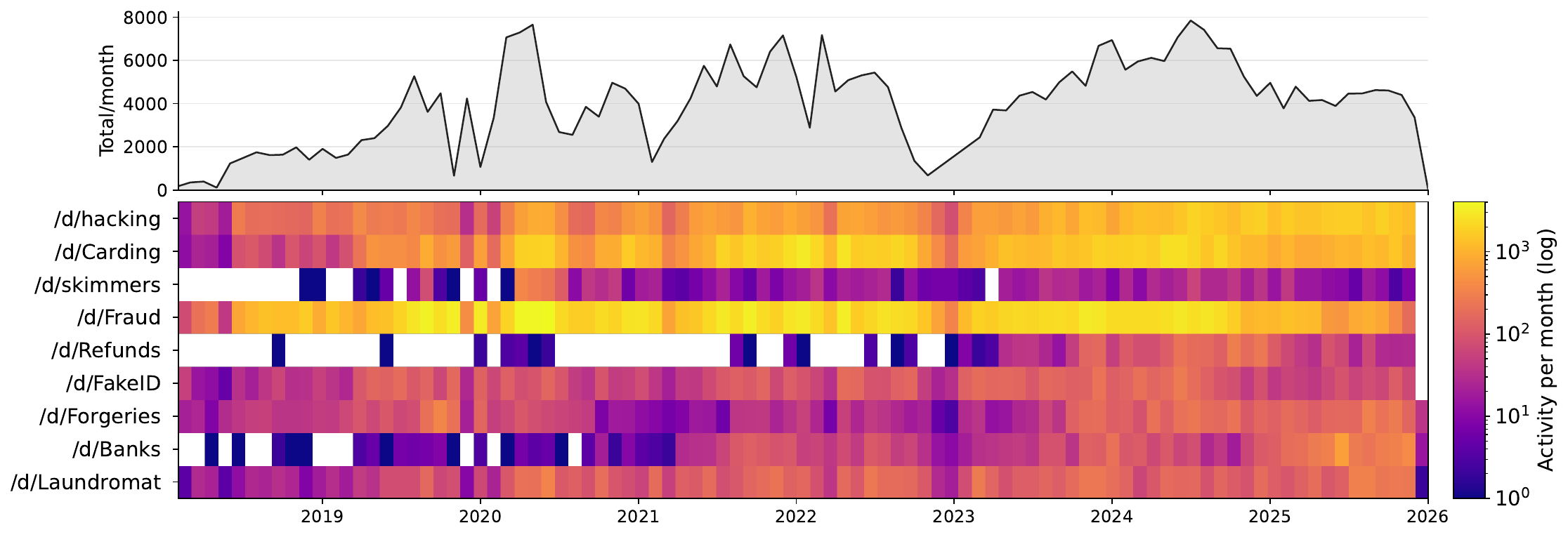}
\caption{Monthly activity across the nine fraud communities, Feb 2018 to Jan 2026. \textbf{Top panel:} Total posts and comments per month across the corpus. \textbf{Bottom panel:} Per-community heatmap with cell color indicating monthly volume on a log scale (white cells indicate zero activity for that month).}
\label{fig:activity_heatmap}
\end{figure*}

\subsection{Data Collection}
\label{sec:collection}

\noindent
Data collection spanned all posts and associated comments published across the nine selected subdreads between February 2018 and January 2026. Following the scraping pipeline outlined by Dong et al. \cite{dong2025parallel}, we routed automated traffic over a local SOCKS5 proxy to the Tor network, paginating through each community's thread index to ensure exhaustive coverage. The data collection was entirely passive, involving no content solicitation or participatory interaction with forum users. Please also refer to our detailed discussion of institutional safeguards provided in the Ethical Considerations section.

\subsection{Analytical Framework}
\label{sec:framework}
\noindent
Rather than treating these subdreads as a flat collection of specialized spaces, we organized them along the operational chain that a fraud scheme typically traverses from preparation to monetization. To characterize these roles, we read each community's stated purpose alongside a sample of its most engaged threads and summarized its dominant function. Drawing on staged models of offending, the crime-script tradition \cite{cornish1994procedural}, the intrusion kill chain \cite{hutchins2011intelligence}, and cybercrime-as-industry research \cite{lusthaus2018industry, holt2013exploring}, we partitioned the nine communities into six categories that serve as the organizing framework for the analyses that follow. We adopt this vocabulary as it closely links our empirical categorization to established literature.


The proposed value chain consists of six non-linear analytical stages:
\begin{itemize} 
    \item \textbf{Capability (Technical Infrastructure):} This initial stage serves as the foundational technical layer of the ecosystem, centered in \sub{hacking}. Discussion here includes acquiring fundamental technical tools and infrastructure, such as malware engineering, server vulnerabilities, botnet configuration, and compromise tools.
    
    \item \textbf{Data Acquisition (Illicit Supply Sourcing):} Once fundamental technical capabilities are established, operations require the raw material of fraud: illicitly obtained financial tokens or data. Represented by \sub{Carding} and \sub{skimmers}, this stage is dedicated to sourcing compromised data or illicit financial instruments through digital or physical channels, including compromised credit card information, hardware-level interception data, and unverified personally identifiable information (PII).
    
    \item \textbf{Technique (Methodological Codification):} This upstream stage focuses on the codification of methodology, ranging from general tradecraft to narrow specializations, represented by \sub{Fraud} and \sub{Refunds}. Topics include executable fraud strategies such as socially engineered chargeback loops or systemic corporate policy workarounds. 
    
    \item \textbf{Identity (Credential Fabrication):} Transitioning toward operational execution, the downstream layer requires structural authentication to deceive defensive systems. Hosted within \sub{FakeID} and \sub{Forgeries}, this stage covers the construction or forgery of credentials needed to effectively utilize acquired data, such as physical passport forgery and advanced KYC (Know Your Customer) documentation circumvention.
    
    \item \textbf{Execution (Targeted Exploitation):} This stage represents the live deployment of upstream methods and fabricated identities against target institutions. Centered in \sub{Banks}, conversation here is intensely operationalized, focusing on current bank security postures, anti-fraud triggers, and remote access trojan (RAT) deployment.
    
    \item \textbf{Extraction (Monetization \& Laundering):} The final stage completes the value chain by converting localized illicit proceeds into permanent, unencumbered, and untraceable wealth. Managed in \sub{Laundromat}, this stage isolates financial laundering pipelines, including cryptocurrency mixing, peer-to-peer cash-out networks, mules management, and digital-to-fiat conversion strategies.
\end{itemize}
These stages serve as an analytical framework; we do not imply that every fraud scheme traverses this continuum linearly.

\subsection{Dataset Overview}
\label{sec:overview}

\noindent
We first summarize the final corpus in Table~\ref{tab:dataset_overview}. Aggregating across all nine subdreads, the dataset comprises 67,976 posts and 305,048 comments authored by 50,761 unique user accounts. These numbers exclude deleted users (\texttt{[deleted]}, \texttt{[removed]}) and the platform's automated moderation account (\texttt{AutoModerator}). The overall volume is dominated by \sub{Fraud}, \sub{Carding}, and \sub{hacking}, which account for the majority of the corpus. While specialized communities such as \sub{Refunds} and \sub{skimmers} contribute smaller but topically focused volumes. 

Fig.~\ref{fig:activity_heatmap} illustrates the longitudinal distribution of monthly activity. Total volume ranges from roughly 250 to 7,500 posts and comments per month, with sustained activity throughout the period and no extended platform-wide silences. The three high-volume upstream communities remain consistently active across the full window, while smaller communities (\sub{skimmers}, \sub{Refunds}, \sub{Banks}) exhibit intermittent quiet months in the early years that resolve into steady activity from 2021 onward.

\section{Findings}
\subsection{Participation and Ecosystem Coupling}
\label{sec:rq1}

\noindent
Our corpus comprises 50,761 distinct user accounts across all nine subdreads. We observe that 8.8\% of users create 48.4\% of all posts and comments across multiple subdreads. At the same time, the remaining 91.2\% contribute briefly to a single community and mostly do not return. We describe these two populations, the structure of their collaboration, and what their distribution means for ecosystem cohesion.

\begin{figure}[t]
\centering
\includegraphics[width=\linewidth]{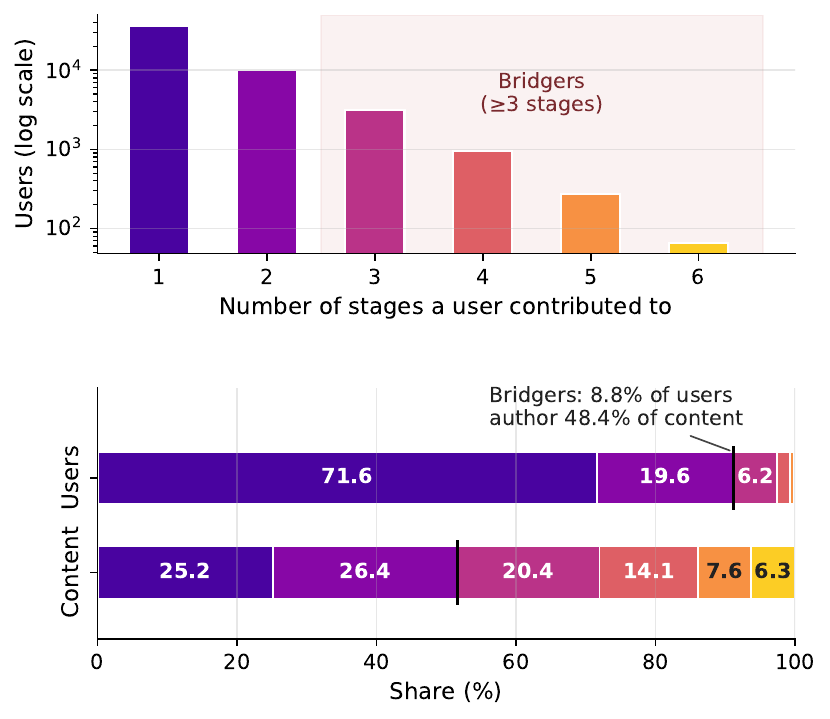}
\caption{Participation breadth and content concentration across stages. \textbf{Top panel:}  Distribution of users by the number of stages they contribute to (log scale); the shaded region marks bridgers, defined as users active in three or more stages. \textbf{Bottom panel:} Share of users (top bar) and share of corpus content (bottom bar) by participation breadth. 71.6\% of users contribute to a single stage, yet the 8.8\% of bridgers produce 48.4\% of all content.}
\label{fig:breadth_and_content}
\end{figure}

\medskip
\noindent
{\textbf{Participation Breadth.}}
Participation breadth, measured as the number of communities to which a user contributes, is sharply heavy-tailed (Fig.~\ref{fig:breadth_and_content}). Of all users, 71.6\% contribute to exactly one of the nine communities, 90.6\% to two or fewer, and only five accounts (0.01\%) appear in all nine. At the stage level, 71.6\% of users are confined to a single stage of the value chain, while only 8.8\% reach three or more stages. The mean breadth is 1.43 communities per user; the median is 1. The typical user is a single-community specialist.


\medskip
\noindent\textbf{Cross-stage Co-occurrence.}
For each pair of stages $(A, B)$, we count the number of users who participate in both. To assess whether this co-occurrence is structurally meaningful rather than an artifact of community sizes and activity heterogeneity, we compare the observed count to its distribution under a \emph{degree-preserving permutation null}. Each user's number of subdreads is held fixed across 500 random reassignments, isolating structural co-occurrence from co-occurrence that activity alone would predict.  We report the resulting \emph{Z-score} $=(\text{observed} - \text{null mean}) / \text{null SD}$. $Z > 0$ indicates that the two stages share users more than chance would predict; $Z < 0$ indicates fewer shared users than chance. Full pair-wise statistics, including observed and expected counts, are reported in Table~\ref{tab:zscores}.

The upstream stages (Capability, Data Acquisition, and Technique) share users at rates far exceeding the null expectation: Acquisition $\times$ Technique overlap is $3.2\times$ the null ($Z=154$), Capability $\times$ Technique is $2.7\times$ ($Z=80$), Capability $\times$ Acquisition is $2.2\times$ ($Z=52$). This disproportionate co-occurrence identifies a tightly coupled \emph{upstream knowledge cluster}. Downstream is the inverse. Identity, Execution, and Extraction co-occur \emph{below} the null model. Execution $\times$ Extraction at $0.29\times$ ($Z=-25$), Identity $\times$ Execution at $0.19\times$ ($Z=-38$), and Identity $\times$ Extraction at $0.30\times$ ($Z=-32$). Cross-stage flow between upstream and downstream is also mostly below null. It suggests that downstream users are more likely to be highly specialized, forming three distinct operator populations rather than a coordinated monetization layer. The overall architecture rejects both conventional readings of the ecosystem (a single, tightly coupled community with a flat federation of independents) in favor of an asymmetric structure: a tightly coupled upstream attached to specialized downstream populations.

\begin{table}[t]
\centering
\caption{Observed versus expected overlap between stage pairs. Positive $Z$ values indicate greater-than-expected overlap, while negative values indicate under-representation.}
\resizebox{\columnwidth}{!}{
\label{tab:zscores}
\begin{tabular}{llccc}
\toprule
\textbf{Stage A} & \textbf{Stage B} & \textbf{Observed} & \textbf{Expected} & \textbf{Z} \\
\midrule
Acquisition & Technique & 9,226 & 2,892.0 & +154.1 \\
Capability & Technique & 4,362 & 1,600.3 & +80.0 \\
Capability & Acquisition & 3,462 & 1,603.1 & +51.9 \\
Capability & Extraction & 985 & 894.7 & +3.4 \\
Technique & Extraction & 1,396 & 1,602.1 & -5.6 \\
Technique & Execution & 1,156 & 1,602.6 & -13.2 \\
Capability & Execution & 539 & 891.6 & -13.3 \\
Acquisition & Extraction & 1,067 & 1,603.0 & -14.5 \\
Technique & Identity & 2,221 & 2,899.0 & -15.8 \\
Acquisition & Execution & 1,022 & 1,602.7 & -16.3 \\
Capability & Identity & 994 & 1,607.1 & -16.3 \\
Execution & Extraction & 259 & 891.6 & -24.5 \\
Identity & Extraction & 481 & 1,605.7 & -31.8 \\
Acquisition & Identity & 1,485 & 2,896.7 & -34.3 \\
Identity & Execution & 298 & 1,602.2 & -37.9 \\
\bottomrule
\end{tabular}}
\end{table}

\medskip
\noindent\textbf{Bridging Population.}
We refer to users active in three or more stages as \emph{bridgers}. The 4,468 bridger accounts (8.8\% of users) author 48.4\% of all corpus content. Their median post-plus-comment count is 15, compared to 2 for non-bridgers ($7.5\times$); their median observed tenure is 206 days, compared to under one day for non-bridgers.

Bridgers differ in which stages they span. One-third (33.2\%; 1484 users) operate exclusively as `knowledge browsers' who engage with all three upstream stages of Capability, Acquisition, and Technique without any downstream activity. Roughly half (49.7\%; 2,221 users) cross from upstream into exactly one downstream stage, primarily targeting Identity (22.3\% of bridgers), Extraction (14.6\%), or Execution (12.8\%). A further 15.0\% (671 users) span two downstream stages. At the extreme, a cohort of 89 \emph{full-chain operators} (0.18\% of all users) is active in all three downstream stages plus at least one upstream stage. With a median tenure exceeding two years, these prolific accounts average a median output of 123 events, nearly an order of magnitude above the typical bridger, and alone generate 6.8\% of the entire corpus.

\subsection{Production and Engagement Distribution}
\label{sec:rq2}

\noindent
At the user level, the ecosystem is asymmetric: a tightly coupled upstream knowledge cluster, three specialized downstream populations, and a thin bridging layer between them. We ask whether the same asymmetry holds at the level of content. The findings reframe the picture in two ways. The upstream/downstream volume gap is substantially sharper than the user-level analysis alone suggested. But once engagement and author-concentration are introduced, the downstream resolves not into a single operational mode but into at least two distinct ones, and one downstream community (Extraction) behaves more like a discussion-heavy upstream stage than a quiet operational venue.

\begin{table}[t]
\centering
\caption{Content volume per value-chain stage. The comments-to-posts ratio measures whether a stage's content is dominated by reply (discursive use) or by origination (transactional use).}
\label{tab:content_volume}
\begin{tabular}{lrrr}
\toprule
\textbf{Stage} & \textbf{Posts} & \textbf{Comments} & \textbf{C/P ratio} \\
\midrule
Capability   & 11,814 & 53,518  & 4.53 \\
Acquisition  & 17,690 & 84,369  & 4.77 \\
Technique    & 31,750 & 139,540 & 4.39 \\
Identity     & 3,620  & 13,025  & 3.60 \\
Execution    & 1,476  & 4,991   & 3.38 \\
Extraction   & 1,626  & 9,605   & 5.91 \\
\midrule
\textbf{Total} & \textbf{67,976} & \textbf{305,048} & \textbf{4.49} \\
\bottomrule
\end{tabular}
\end{table}

\medskip
\noindent\textbf{Volume and Structure of Content.}
The three upstream stages produce 90.8\% of all corpus content (Table~\ref{tab:content_volume}). Technique alone accounts for 45.9\% of the 373{,}024 posts and comments in the corpus; the three downstream stages together contribute 9.2\%. Controlling for the number of communities at each end of the chain, the upstream's share of activity is roughly an order of magnitude greater than the downstream's per-community average.

The comments-to-posts (C/P) ratio is broadly similar across stages (3.4-5.9) and does not, on its own, distinguish upstream from downstream content: pooled ratios are 4.56 upstream and 4.30 downstream, a difference of only 1.1$\times$. The variation that does exist runs against the simple two-tier prediction. Identity (3.60) and Execution (3.38) sit at the low end, consistent with venues whose posts function as listings or announcements rather than as prompts for discussion. But Extraction sits at the highest comments-to-posts ratio of any stage (5.91), indicating that despite its small absolute volume, each Extraction post triggers more reply activity per origination than any post elsewhere in the corpus. Whatever ``downstream'' means as a content category, Extraction does not belong to it without qualification.

\begin{table}[t]
\centering
\caption{Pairwise comparisons of comments-per-post distributions. Values represent Mann-Whitney $U$ tests and rank-biserial $r$ effect sizes, where $r > 0$ denotes that Stage A receives a higher volume of comments per post than Stage B.} 
\label{tab:engagement_pairwise}
\resizebox{\columnwidth}{!}{
\begin{tabular}{llrrr}
\toprule
\textbf{Stage A} & \textbf{Stage B} & \textbf{Median A} & \textbf{Median B} & \textbf{$r$} \\
\midrule
\rowcolor{lightblue}\multicolumn{5}{l}{\emph{Extraction outperforms every other stage}} \\
Extraction  & Execution    & 4 & 2 & +0.34$^{**}$ \\
Extraction  & Identity     & 4 & 2 & +0.33$^{**}$ \\
Extraction  & Technique    & 4 & 3 & +0.21$^{**}$ \\
Extraction  & Capability   & 4 & 3 & +0.16$^{**}$ \\
Extraction  & Acquisition  & 4 & 3 & +0.16$^{**}$ \\
\midrule
\rowcolor{lightblue}\multicolumn{5}{l}{\emph{Upstream outperforms Identity and Execution}} \\
Capability  & Execution    & 3 & 2 & +0.20$^{**}$ \\
Capability  & Identity     & 3 & 2 & +0.19$^{**}$ \\
Acquisition & Execution    & 3 & 2 & +0.19$^{**}$\\
Acquisition & Identity     & 3 & 2 & +0.18$^{**}$ \\
Technique   & Execution    & 3 & 2 & +0.14$^{**}$ \\
Technique   & Identity     & 3 & 2 & +0.13$^{**}$ \\
\midrule
\rowcolor{lightblue}\multicolumn{5}{l}{\emph{Within-cluster contrasts: near-equivalent}} \\
Capability  & Technique    & 3 & 3 & +0.06$^{**}$ \\
Acquisition & Technique    & 3 & 3 & +0.05$^{**}$ \\
Capability  & Acquisition  & 3 & 3 & +0.01 \\
Identity    & Execution    & 2 & 2 & +0.01 \\
\bottomrule
\multicolumn{5}{l}{\small $^{**}$ marks significance of $p < 0.05$} \\
\end{tabular}}
\end{table}

\medskip
\noindent{\textbf{Engagement Across Stages.}}
The volume measures describe what is produced. To characterize how content is \emph{received}, we report the median number of comments per post within each stage and compare distributions pairwise using the Mann-Whitney $U$ test. Effect sizes are reported as rank-biserial $r$, consistent with the previous findings; $r > 0$ indicates Stage A receives more comments per post than Stage B.

The pattern is not the binary upstream-vs-downstream gradient that the user-level analysis would predict. Three distinct engagement regimes emerge (Table~\ref{tab:engagement_pairwise}). The upstream stages cluster at a median of 3 comments per post and are statistically near-equivalent: Capability vs.\ Acquisition is null ($r = +0.01$, $p = 0.30$), and the within-upstream effects against Technique remain below $r = 0.06$. Identity and Execution sit \emph{below} the upstream cluster, with a median of 2 comments per post, and effects of $+0.13$ to $+0.20$ against each upstream stage; modest but consistent. Extraction sits \emph{above} all other stages at a median of 4, attracting more comments per post than even Capability (Extraction vs.\ Capability $r = +0.16$, $p < 0.001$) and substantially more than the other two downstream stages (Extraction vs.\ Execution $r = +0.34$).

\begin{figure*}[t]
\centering
\includegraphics[width=\textwidth]{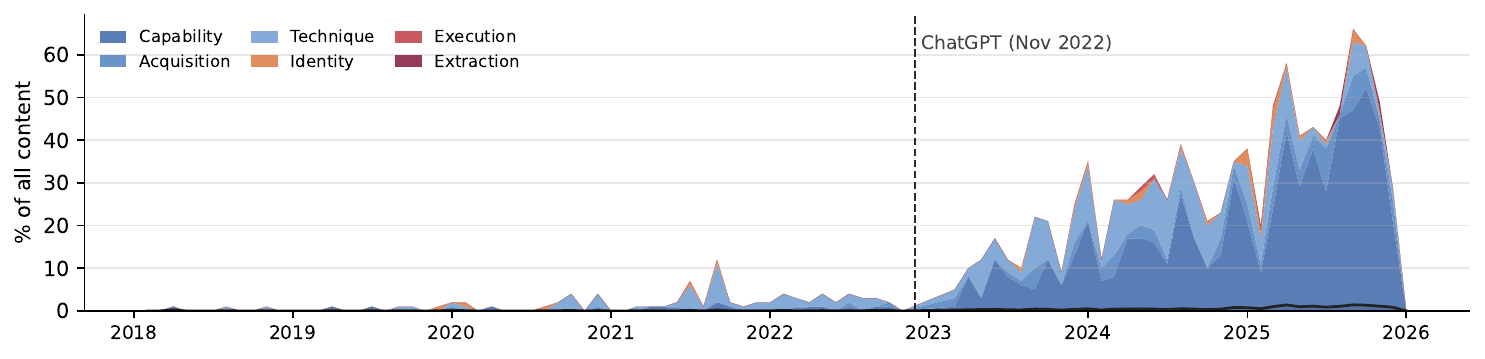}
\caption{Longitudinal diffusion of generative AI mentions across the fraud value chain.  The vertical dashed line marks the public release of ChatGPT (November 30, 2022), illustrating the inflection point between baseline discussion and widespread technological adoption.}
\label{fig:ai_temporal}
\end{figure*}

Two findings follow. First, the downstream is not a uniform operational regime. Identity and Execution receive limited reply activity per post, consistent with venues whose primary outputs are listings, offers, and short transactional exchanges; their downstream consequences presumably occur off-platform via direct messages or escrow channels. Extraction, by contrast, behaves like a discussion-heavy specialty community: laundering and cash-out workflows are sufficiently complex that posts attract sustained replies, troubleshooting, and follow-up across multiple commenters. Second, the upstream cluster is internally homogeneous on this metric: Capability, Acquisition, and Technique are interchangeable to within rank-biserial effects of $0.06$. The user-level upstream cluster identified in Section~\ref{sec:rq1} has a content-level counterpart, but the downstream does not.

\medskip
\noindent{\textbf{Author Concentration within Stages.}}
Section~\ref{sec:rq1} reported that 8.8\% of users produce 48.4\% of the corpus content. Within stages, this concentration is not uniform. Contrary to the prediction that downstream stages, being narrower in absolute participation, would also be more concentrated, the upstream stages exhibit \emph{higher} author concentration than the downstream stages (Table~\ref{tab:author_concentration}).

The Gini coefficient over events-per-author rises monotonically across the upstream cluster, $0.61$ in Capability, $0.65$ in Acquisition, $0.70$ in Technique, and falls in the downstream stages to $0.55$ in Identity, $0.55$ in Extraction, and $0.52$ in Execution. The top-10\% share moves in the same direction: 62.5\% in Technique against 45.8\% in Execution. Although the upstream stages have far more authors in absolute terms, 25,638 in Technique against 2,065 in Execution, a smaller proportion of those authors carry most of the load. Technique looks like a power-law knowledge community: a long tail of occasional commenters surrounding a small expert core that produces the substantive content. Execution, by contrast, looks like a small but comparatively egalitarian operator community where most participants contribute at broadly similar levels.

This finding is consistent with the bridger pattern documented before: the 8.8\% of corpus-wide bridgers who generate 48.4\% of all content are disproportionately clustered within the upstream expert cores identified by the high within-stage Gini coefficients. Conversely, the downstream stages exhibit flatter author distributions and fewer per capita heavy-hitters, indicating an egalitarian population of uniformly active operators rather than a dominant elite atop a wide lurker base.

\begin{table}[t]
\centering
\caption{Within-stage author concentration. Higher Gini and higher top-10\% share both indicate that fewer authors produce a greater share of the stage's content.}
\label{tab:author_concentration}
\begin{tabular}{lrrr}
\toprule
\textbf{Stage} & \textbf{Authors} & \textbf{Gini} & \textbf{Top-10\% share} \\
\midrule
Capability   & 15,025 & 0.61 & 53.8\% \\
Acquisition  & 19,720 & 0.65 & 57.5\% \\
Technique    & 25,638 & 0.70 & 62.5\% \\
Identity     & 5,426  & 0.55 & 50.4\% \\
Execution    & 2,065  & 0.52 & 45.8\% \\
Extraction   & 3,478  & 0.55 & 48.7\% \\
\bottomrule
\end{tabular}
\end{table}

\subsection{Generative AI Footprint}
\label{sec:rq3}

\noindent
The fraud ecosystem we have characterized predates the public release of generative AI, but the November 2022 release of ChatGPT introduced capabilities relevant to Fraud at scale: text generation for phishing and social-engineering scripts, image and voice synthesis for identity fabrication, and conversational interfaces that lower the technical barrier to operating attacks. To assess where in the chain these capabilities are being adopted, we curated a vocabulary of AI-related terms and scanned the corpus for matches (see Fig.~\ref{fig:ai_temporal}).

\medskip
\noindent{\textbf{Vocabulary and Matching.}}
We compiled 32 AI-related expressions grouped into five categories: General-Purpose LLMs (including ChatGPT, GPT-$x$ variants, Claude, etc.), Synthetic Media \& Deepfakes (including Stable Diffusion, DALL-E, ElevenLabs), Adversarial \& Malicius LLMs (WormGPT, FraudGPT), Prompt Engineering (generative AI, prompt jailbreaks, AI bypass), and AI Corporate Entities (OpenAI, Anthropic). Table~\ref{tab:ai_terms} lists all the regular expressions used to detect AI-related content in the corpus. Each expression is a case-insensitive regular expression with word boundaries; ambiguous terms were narrowed to require contextual prefixes (e.g., `github copilot' rather than the bare `copilot'). We validated the vocabulary by inspecting matched content samples for each term.

\begin{table*}[h]
\centering
\caption{Taxonomy of the generative AI vocabulary and category-level match totals. Term keys are ordered by descending frequency. Total count (marked *) exceed the 1,102 unique AI-related events due to co-occurring terms within a single post or comment.}
\label{tab:ai_terms}
\setlength{\tabcolsep}{4pt}
\renewcommand{\arraystretch}{1.2}

\begin{tabular}{p{4cm}p{12cm}r}
\toprule
\textbf{Category} & \textbf{Matched Terms} & \textbf{Matches} \\
\midrule
\textsf{General-Purpose LLMs}
 & ChatGPT, LLM, large language model, GPT, Llama, Ollama, Grok, Claude, GPT-2/3/4, Google Bard, Gemini, GitHub/Microsoft Copilot, Vicuna, OpenAI/GitHub Codex
 & 968 \\
\textsf{Synthetic Media \& Deepfakes}
 & deepfake, DeepFaceLab, FaceSwap, Stable Diffusion, OnlyFakes, voice cloning, DALL-E, ElevenLabs, NightCafe, synthetic voice/media, MidJourney
 & 242 \\
\textsf{Adversarial \& Malicius LLMs}
 & WormGPT, FraudGPT, EvilGPT, DarkGPT, WolfGPT
 & 109 \\
\textsf{Prompt Engineering}
 & 
 prompt engineering, prompt injection, AI bypass, KYC verification, prompt jailbreak, DAN, 
 custom GPT
 & 77 \\
\textsf{AI Corporate Entities}
 & Hugging Face, OpenAI, Anthropic
 & 62 \\
\midrule
\textbf{Total} & & \textbf{1,458*} \\
\bottomrule
\end{tabular}
\end{table*}

Across the corpus, 271 posts (0.40\%) and 831 comments (0.27\%) match at least one AI term, for a total of 1,102 AI-related events (0.30\% of all content). Three terms dominate the vocabulary: ChatGPT (375 matches), LLM (219), and deepfake (218). Notably, fraud-specific LLMs such as WormGPT, FraudGPT, EvilGPT, DarkGPT, and WolfGPT appear 109 times, indicating that purpose-built AI tools for cybercrime have a measurable presence in the ecosystem.

\medskip
\noindent{\textbf{AI Discussion Concentrates Upstream.}}
Table \ref{tab:ai_per_stage} shows the distribution of AI-matched content across the six value-chain stages. Capability dominates among all the stages, with \sub{hacking} alone accounting for 682 of 1,102 AI events (62\% of all AI discussion in the corpus), and AI terms appear in 1.04\% of its content, roughly six times the share in any other stage. Technique follows at 0.17\% (286 events), almost entirely from \sub{Fraud}. Acquisition trails at 0.09\%. The three downstream stages register negligible AI activity: Identity at 0.18\%, Execution at 0.05\%, Extraction at 0.04\%.

The pattern mirrors the upstream and downstream asymmetry observed in Section \ref{sec:rq1}, but in a more concentrated form: AI discussion is not merely upstream-heavy; it is overwhelmingly a Capability phenomenon. Even within the upstream knowledge cluster, where ordinary content is roughly balanced across \sub{hacking}, \sub{Carding}, and \sub{Fraud}, AI discussion is heavily skewed toward \sub{hacking}. This positions Capability as the ecosystem's principal site of AI uptake, where new tooling is first encountered, evaluated, and repurposed.

\medskip
\noindent{\textbf{ChatGPT Release.}}
The AI vocabulary returns matches as early as April 2018, well before generative AI entered mainstream use, as shown in Fig.~\ref{fig:ai_temporal}. To distinguish the background discussion of machine learning from the post-ChatGPT surge, we partition AI-matched content around the public release of ChatGPT on 30 November 2022. The 58 months preceding the release contain only 83 AI mentions (1.4 per month); the 35 months following it contain 1,019 events at 29.1 per month, a 20.3$\times$ increase in monthly rate. Put differently, 92.5\% of all AI-related content in our corpus arrived after ChatGPT, even though that window covers only 38\% of the observed timeline.

Table \ref{tab:ai_pre_post} breaks the shift down by stage. The increase is not uniform. Capability's monthly rate rises by a factor of 79.1, Acquisition by 24.9, and Identity by 8.3. Technique shows the smallest relative jump (6.5$\times$) because it carries the largest pre-existing baseline: 58 of its 286 total events predate ChatGPT, reflecting discussion of machine-learning methods and statistical techniques that predate generative AI. The first observed AI mention in Capability is dated April 2018, and in Technique August 2018, confirming that the upstream stages already hosted some AI-adjacent vocabulary before the inflection.

The two downstream stages tell a different story. Execution and Extraction contains \emph{no} AI mentions whatsoever in the 58 months preceding ChatGPT. The first AI-related Execution event appears in May 2024, 18 months after ChatGPT's release, and the first Extraction mention in February 2025, 27 months after. These stages do not exhibit a pre/post-shift rate changes; they exhibit a delayed onset. Whatever AI uptake exists at the downstream end of the value chain is not an acceleration of prior interest but the introduction of a topic with no prior representation at all, and even then, at negligible absolute volume (3 and 5 events, respectively).

Generative AI did not enter the fraud value chain at all sites simultaneously. Capability absorbed it immediately and at scale, building on a small but established baseline of ML-adjacent discussion; Acquisition and Identity followed at lower intensity; Execution and Extraction registered the technology only after substantial lag, once specific operational use cases became salient. The diffusion of generative AI across the ecosystem is therefore concentrated upstream, both in volume and in timing.

\begin{table}[t]
\centering
\caption{AI-matched content per value-chain stage. AI events = posts + 
comments mentioning any term from the AI vocabulary.}
\label{tab:ai_per_stage}
\begin{tabular}{lrr}
\toprule
\textbf{Stage} & \textbf{AI events} & \textbf{\% of stage} \\
\midrule
Capability    &   682 & 1.04\% \\
Technique     &   286 & 0.17\% \\
Identity      &    30 & 0.18\% \\
Acquisition   &    96 & 0.09\% \\
Execution     &     3 & 0.05\% \\
Extraction    &     5 & 0.04\% \\
\midrule
\textbf{Total} & \textbf{1,102} & \textbf{0.30\%} \\
\bottomrule
\end{tabular}
\end{table}

\medskip
\noindent{\textbf{Engagement Scale.}}
Volume measures the \emph{presence} of AI in the corpus, but not its \emph{reception}. A topic that occupies less than 1\% of the content could still command disproportionate attention, or it could be largely ignored. We test this by comparing comment counts on AI-tagged posts with those on non-AI posts within the same stage, restricting the analysis to posts published after the release of ChatGPT so that both groups have had comparable time to accumulate replies. Per-stage differences are evaluated using the Mann-Whitney $U$ test; we report rank-biserial $r$ as the effect size, where $r > 0$ indicates that AI-tagged posts attract more comments than non-AI posts within the same stage.

\begin{table}[t]
\centering
\caption{AI mentions per stage, before and after the public release of ChatGPT (30 November 2022). Monthly rates normalize for the 58-month pre- and 35-month post-release windows.}
\label{tab:ai_pre_post}
\begin{tabular}{lrrrrr}
\toprule
\textbf{Stage} & \textbf{Pre} & \textbf{Post} & \textbf{Pre/mo} & \textbf{Post/mo} & \textbf{Ratio} \\
\midrule
Capability    & 14 & 668 & 0.2 & 19.1 & 79.1$\times$ \\
Acquisition   &  6 &  90 & 0.1 &  2.6 & 24.9$\times$ \\
Technique     & 58 & 228 & 1.0 &  6.5 &  6.5$\times$ \\
Identity      &  5 &  25 & 0.1 &  0.7 &  8.3$\times$ \\
Execution     &  0 &   3 & 0.0 &  0.1 & --- \\
Extraction    &  0 &   5 & 0.0 &  0.1 & --- \\
\midrule
\textbf{Total} & \textbf{83} & \textbf{1,019} & \textbf{1.4} & \textbf{29.1} & \textbf{20.3$\times$} \\
\bottomrule
\end{tabular}
\end{table}

\begin{figure}[t]
  \centering
  \includegraphics[width=\columnwidth]{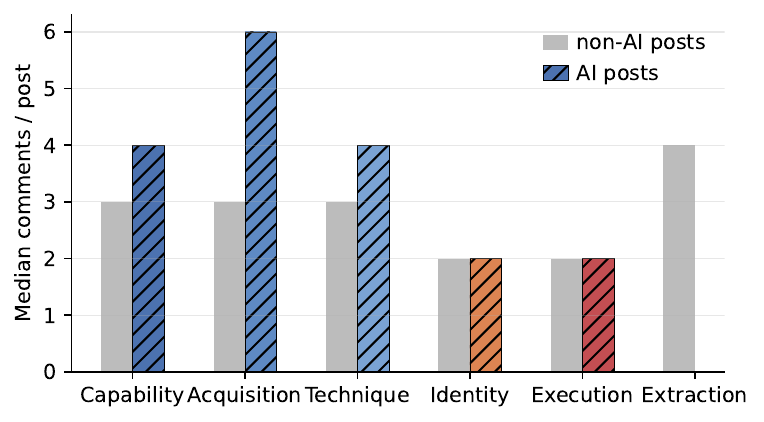}
  \caption{Median comments per post per value-chain stage, AI-tagged (dashed) vs non-AI (gray), after the public release of ChatGPT (30 November, 202). Extraction contains no AI-tagged posts in this period.}
  \label{fig:ai_engagement}
\end{figure}

Fig.~\ref{fig:ai_engagement} shows the comparison. The pooled effect is small but positive ($r = +0.13$, $p < 0.001$, $n_{\mathrm{AI}} = 232$) and decomposes asymmetrically across stages. In Capability, the stage hosting the majority of AI discussion, the effect is null ($r = +0.03$, $p = 0.52$, $n_{\mathrm{AI}} = 135$): AI threads attract engagement indistinguishable from that of other threads. In Technique, the effect is small but significant ($r = +0.18$, $p = 0.02$, $n_{\mathrm{AI}} = 60$). In Acquisition, where AI mentions are rare (29 AI-tagged posts among $\sim$7,000 total posts), AI threads receive a median of 8 comments versus 4 for non-AI threads in the same stage, corresponding to a medium-to-large effect ($r = +0.37$, $p < 0.001$). Identity, Execution, and Extraction each contain six or fewer AI-tagged posts in the post-release window and are therefore omitted from the comparison.

The engagement premium scales inversely with the level of AI discussion saturation within a stage. This pattern is more consistent with a \emph{novelty} response than with a substantive shift in how the ecosystem produces and consumes fraud content: users engage with AI as remarkable where it remains unusual, but treat it as routine in Capability, where it has become part of the ambient discourse. The post-ChatGPT surge is real and is overwhelmingly concentrated upstream, but within the upstream stages themselves, AI content does not receive disproportionate attention. AI has become \emph{frequent} in Capability without becoming \emph{exceptional}.

\section{Discussion}
\label{sec:discussion}

\noindent
So far, we have characterized the underground fraud ecosystem as a distinctive geometric system. At the user level, it is held together by a small connective population that spans otherwise specialized communities. At the content level, it resolves into a two-tier upstream/downstream split that performs different functions and absorbs different kinds of activity. At the level of technological diffusion, generative AI is entering through the same asymmetric channel that the user and content structures imply, upstream first, downstream barely. We now discuss these findings and what they imply for how the ecosystem functions, how it is likely to absorb future technological shocks, and where intervention might be directed.

\subsection{Asymmetric Ecosystem}

\noindent
The most significant finding of \textbf{RQ1} is participation asymmetry. Among 50,761 accounts, 71.6\% operate exclusively within a single stage of the fraud value chain, and 90.6\% engage with no more than two communities. Conversely, network integration relies on a critical elite: the 8.8\% bridging population generates 48.4\% of all ecosystem content, while a highly concentrated 89 full-chain operators (0.18\% of users) contribute 6.8\% of the total corpus. Structural continuity is therefore sustained not by broad, distributed cross-stage labor mobility, but by a precise, high-output connective core. These bridging actors exhibit extreme longitudinal persistence, with a median observed lifespan exceeding 200 days, contrasted sharply against the transient participation of non-bridgers, whose median activity window falls below a single day.

Pairwise overlap analysis further clarifies social topology at the individual level. The three upstream phases (i.e., Capability, Acquisition, and Technique) share users at rates two to three times higher than the activity-matched null model, establishing that they operate as a single, highly integrated knowledge cluster rather than distinct communities. Conversely, user co-occurrence across the three downstream stages (i.e., Identity, Execution, and Extraction) falls significantly below the null expectation. This indicates that downstream operators are highly specialized, executing monetization workflows within discrete, insulated populations with minimal personnel overlap, rather than a coordinated operational layer.

Content-level metrics in \textbf{RQ2} reveal an even more pronounced structural asymmetry, suggesting a granular taxonomy that escapes simple binary classification. Several patterns are noticeable. The upstream cluster dominates material production, generating 90.8\% of all corpus content, with the Technique stage alone driving roughly half of the total volume.
Productivity is also heavily stratified. Within-stage author concentration scales significantly higher upstream, with Gini coefficients ranging from 0.61 (Capability) to 0.70 (Technique), contrasted against a tighter 0.52 to 0.55 distribution across the downstream stages. Indeed, the top 10\% of upstream contributors produce more than half of their respective stages' content (53.8\% in Capability, 62.5\% in Technique), compared to a flatter 45.8\% to 50.4\% distribution downstream. The upstream, therefore, reflects the classic dynamics of a power-law knowledge community, where a massive, transient tail surrounds a highly concentrated expert core, whereas downstream spaces operate as small, egalitarian networks without a dominant contributor layer.

However, engagement showed different dynamics. While posts in the Identity and Execution stages function largely as quiet transactional listings with few replies, the Extraction stage emerges as a major conversational anomaly, triggering the highest comment-per-post volume in the entire corpus.


\subsection{Generative AI's Diffusion}

\noindent
Our findings on \textbf{RQ3} shows that generative AI has been absorbed into this ecosystem, mirroring and sharpening its existing geometry. Of all AI-related content, 62\% sits inside a single upstream subcommunity (\sub{hacking}), and 86\% lies in upstream stages overall. The downstream stages register negligible AI activity: Execution and Extraction together contain only 8 AI events out of more than 1,000.

The asymmetry is not only in volume, but also in timing. Prior to the release of ChatGPT in November 2022, Capability and Technique already included AI-related vocabulary, indicating prior interest in AI. Execution and Extraction make no reference to AI in the 58 months leading up to ChatGPT; their first AI occurrences appear 18 and 27 months later. Downstream uptake is not the amplification of earlier interest, but rather the delayed introduction of a previously absent topic. This pattern has two implications. First, our findings only partially corroborate the prevalent narrative about the role of generative AI in cybercrime, which is that it reduces barriers and democratizes attack capabilities throughout the threat environment.
Second, the inverse relationship between engagement with AI and AI saturation is informative about how the upstream is integrating these tools. In Capability, the stage hosting the majority of AI discussion, AI-tagged posts receive engagement statistically indistinguishable from non-AI posts. In Acquisition, where AI mentions are rare, AI-tagged posts attract a medium-to-large engagement premium. This is consistent with a novelty response. AI is remarkable where it remains unusual, but treated as routine where it has become ambient. Capability has naturalized generative AI; it is part of the discourse rather than a topic within it. This naturalization is itself a finding, as AI is being integrated into existing capability-building practices rather than reshaping them.

\subsection{Extraction Anomaly}

\noindent
By absolute volume (11,231 events) and author count (3,478 distinct users), Extraction belongs to the operational downstream. But by engagement profile, it aligns with upstream discussion communities. Each Extraction post attracts a median of 4 comments, against 3 in the upstream stages and 2 in Identity and Execution. Its comments-to-posts ratio (5.91) exceeds every other stage in the corpus, including Capability's substantial 4.53.

Several explanations are viable. Laundering and cash-out workflows are inherently case-specific, requiring a unique combination of source funds, target jurisdiction, available conversion tools, and tolerance for delays. Posts requesting such scenarios may necessitate elaboration and follow-up that a generic carding tutorial does not. Second, small-community effects: with fewer than 3,500 distinct writers, Extraction may be small enough to support strong mutual replying norms. Third, the real repercussions of an extraction error, such as lost funds, frozen accounts, or traceable transactions, may necessitate a more careful peer review than a similar Identity or Execution post.

We cannot distinguish between these explanations from the present corpus alone; doing so would require in-depth analysis of the Extraction thread content. We flag the finding here as an open question and note that, whichever explanation holds, Extraction is the only downstream stage that combines operational substance with active community discussion. If generative AI ever finds substantial operational use in this ecosystem, Extraction is the stage we would expect to see it surface in first.

\subsection{Implications}
\noindent
Our findings argue for several reframings of how the underground fraud ecosystem should be analyzed and acted upon.

\medskip
\noindent{\textbf{For platform and enforcement actors,}} the locus of disruption is not the long tail. Excluding the 91\% of single-stage, single-visit users would have a negligible effect on productive output. Leverage could lie with the bridgers, particularly the 89 full-chain operators, whose combined output exceeds that of tens of thousands of casual users. These accounts are also observationally tractable: median tenure exceeds two years, and activity spans multiple communities. Targeting the top decile of upstream contributors would address 54-63\% of stage content while affecting a population two orders of magnitude smaller than the full participant base.

\medskip
\noindent\textbf{For threat intelligence,} the upstream vs.\ downstream split implies that emerging capabilities, AI-enabled phishing, deepfake tooling, and prompt-jailbreak techniques will surface in Capability and Technique long before they reach Execution or Extraction, if at all. Upstream monitoring provides early warning; downstream monitoring observes operational consequences at a substantial lag, with the 18- and 27-month delays in AI reaching the downstream offering a preliminary empirical estimate. Minimizing this lag requires enhancing timely, secure, and lawful information-sharing across institutional boundaries while responsibly deploying emerging AI toolkits to facilitate the proactive investigation and prosecution of these fraud networks.

\medskip
\noindent\textbf{For research on AI-enabled cybercrime,} our findings temper the strongest formulations of the AI-democratization thesis. Fraud-specific LLMs (WormGPT, FraudGPT, EvilGPT, DarkGPT, WolfGPT) account for 109 corpus mentions, yet their footprint remains overwhelmingly upstream and discursive. We find no evidence that the operational downstream has been substantively transformed by these tools. Whether this reflects a genuine ceiling, an unresolved diffusion lag, or an artifact of how operators discuss their work in semi-public forums cannot be adjudicated from this corpus alone.

\subsection{Future Directions}
\noindent Our current study is limited to a single underground forum, Dread, and specifically to a one-hop cluster seeded from \sub{Fraud}. Although this design captures an analytically tractable slice of the fraud ecosystem, it does not exhaust the broader network of marketplaces, encrypted channels, regional-language communities, and cross-platform spaces through which illicit knowledge circulates. Future work should therefore test whether our six-stage structure generalizes across alternative communication infrastructures. This would allow the present framework to be evaluated not only as a description of one Dread-centered ecosystem but also as a broader model of how fraud capabilities, data sources, operational techniques, identity resources, execution strategies, and extraction methods are organized across the underground economy.

More importantly, our framework can be extended from retrospective measurement to prospective early warning. Because emerging capabilities consistently surface in upstream knowledge spaces, especially Capability and Technique, before migrating downstream to operational venues like Execution or Extraction, future systems can utilize the six-stage value chain as an organizing ontology for AI-assisted monitoring. This enables defensive systems to automatically classify external venues, track long-tenured bridgers, and empirically estimate the lead time between upstream conceptualization and downstream operational adoption. In this setting, lead time shifts from an informal judgment to a measurable intelligence quantity. Beyond threat tracking, analyzing the corpus of these critical contributors provides the empirical foundation to design targeted, multi-stakeholder educational campaigns that raise public awareness and build systemic resilience.

A particularly promising direction is the development of agentic threat-intelligence frameworks for passive, human-in-the-loop source discovery. Such systems would not merely collect larger volumes of dark-web content. Instead, bounded AI agents could maintain an evolving intelligence graph of communities, users, aliases, links, services, terms, and techniques; use the six-stage ontology to classify newly observed artifacts; and propose candidate sources when repeated cross-references, user movement, or topic diffusion suggest that an external venue is becoming relevant. This systematically transforms source selection into an explicit, evaluable research problem.

Such an agentic architecture addresses the primary bottleneck in underground monitoring: the cognitive burden and safety risks of sustained manual inspection. Human analysts must spend extensive periods filtering noisy, hostile, and potentially harmful environments to isolate durable signals from transient chatter. Bounded AI agents can absorb this first-pass triage burden by passively processing lawfully accessible material, preserving data provenance, redacting sensitive artifacts, and surfacing only the entities that breach pre-specified intelligence thresholds. The goal is not to replace human judgment or to allow autonomous systems to participate in underground communities, but to minimize analyst exposure while making source discovery more systematic, reproducible, and scalable.

Future evaluations should therefore measure these systems not only by classification accuracy but also by their intelligence value: whether they accelerate high-signal source discovery, improve structural coverage across under-observed operational stages, correctly identify bridgers and expert cores, and provide actionable lead time prior to downstream exploitation. In this sense, the natural next step beyond the static ecosystem map presented here is a continuously updated, agent-assisted model of how fraud capabilities diffuse across underground communities over time.

\section{Ethical Considerations}
\label{sec:ethics}  
\noindent
This study analyzed publicly accessible discussion content from a darknet forum. It was conducted in accordance with the general ethics and DICE-E framework for darknet research \cite{benjamin2019dice} for observational social media research. We summarize below the principal safeguards we have put in place to limit harm to users, third parties, and the research community.

\medskip
\noindent\textbf{Data collection.} The data consists exclusively of post text, comment text, timestamps, pseudonymous usernames, and community membership, all obtained from publicly accessible portions of the forum. No private messages, transaction data, financial credentials, victim-identifying content, or off-platform identifiers were collected or sought. Researchers did not create accounts, post content, or interact with users in any form. The corpus is stored on an encrypted institutional infrastructure with access restricted to the research team.

\medskip
\noindent\textbf{User privacy.} Although forum usernames are pseudonymous, prolific contributors, particularly the bridger population described in  RQ1, could be re-identified through cross-platform correlation. We have therefore reported all findings at the aggregate level: no individual usernames, distinctive verbatim quotations, or identifying linguistic patterns appear in this work. The 89 full-chain operators discussed in this work are characterized solely by aggregate behavioral metrics; no individual accounts are named, quoted, or attributed.

\medskip
\noindent\textbf{Dual-use considerations.} Structural findings about an underground ecosystem can inform both defensive and adversarial actors. We have sought to mitigate this risk by reporting at the level of ecosystem geometry rather than operational practice: the paper describes where generative AI is being adopted and which subpopulations are load-bearing, but does not catalog specific tools, exploitation pipelines, or attack techniques. The fraud-specific LLMs referenced in Section RQ3 (WormGPT, FraudGPT, EvilGPT, DarkGPT, WolfGPT) are already well-documented within academic and threat-intelligence literature; this study introduces no novel offensive capabilities.

\medskip
\noindent\textbf{Disclosure and data sharing.} Aggregate findings will be shared with relevant computer security and law enforcement contacts in advance of publication. To prevent weaponization or widespread exposure, the raw corpus, user-level identifiers, and community-specific operational contents will not be publicly released, though requests for replication will be evaluated case-by-case under strict institutional data-sharing protocols.






\section{Conclusion}
\label{sec:conclusion}
\noindent
Contemporary empirical assessments demonstrate that fraud has evolved into a rapidly shifting, highly sophisticated transnational organized crime. This phenomenon directly threatens individual psychological well-being, macro-economic stability, institutional trust in global financial systems, and broader collective security. We characterized the underground fraud ecosystem along a six-stage value chain, spanning capability development to value extraction. Across user participation, content production, and the diffusion of generative AI, a single picture emerges: a structurally asymmetric ecosystem in which a thin layer of bridgers, concentrated within upstream knowledge cores, carries the bulk of productive output. Generative AI has so far entered along this same channel rather than redrawing it. The leverage points for understanding and intervening in this ecosystem are therefore structural; they lie in the small population of users who connect its stages, not in the tools they are beginning to adopt.

\bibliographystyle{IEEEtran}
\bibliography{bib}

\end{document}